\documentclass{article} % For LaTeX2e
\usepackage{mlgenx_conference,times}

\usepackage{amsmath,amsfonts,bm}

\def\eqref#1{equation~\ref{#1}}
\def\1{\bm{1}}

\DeclareMathAlphabet{\mathsfit}{\encodingdefault}{\sfdefault}{m}{sl}
\SetMathAlphabet{\mathsfit}{bold}{\encodingdefault}{\sfdefault}{bx}{n}

\usepackage{svg}
\usepackage{hyperref}
\usepackage{url}
\usepackage{graphicx}
\usepackage{amsmath,amssymb}
\usepackage{subcaption}
\usepackage{wrapfig}
\usepackage{placeins}  % in preamble

\mlgenxfinal % Uncomment for camera-ready version, but NOT for submission.
\title{Distilling Genomic Models for Efficient mRNA Representation Learning via Embedding Matching}

\author{Rasched Haidari, Sam Martin \& Maxime Allard\textsuperscript{*}
\\
Helical\\
London, UK\\
\texttt{\{maxime\}@helical-ai.com} \\
\textsuperscript{*}Corresponding Author
}

\begin{document}

\maketitle

\begin{abstract}
Large Genomic Foundation Models have recently achieved remarkable results and in-vivo translation capabilities. However these models quickly grow to over a few Billion of parameters and are expensive to run when compute is limited. To overcome this challenge, we present a distillation framework for transferring mRNA representations from a state of the art genomic foundation model into a much smaller model specialized for mRNA sequences, reducing the size by 200-fold. Embedding-level distillation worked better than logit based methods, which we found unstable. Benchmarking on mRNA-bench demonstrates that the distilled model achieves state-of-the-art performance among models of comparable size and competes with larger architectures for mRNA-related tasks. Our results highlight embedding-based distillation of mRNA sequences as an effective training strategy for biological foundation models. This enables similar efficient and scalable sequence modelling in genomics, particularly when large models are computationally challenging or infeasible.
% The abstract must be limited to one
% paragraph. If you use the provided LaTex template ant style files, you will successfully adhere to the following formatting requirements: The abstract paragraph should be indented 1/2~inch (3~picas) on both left and
% right-hand margins. Use 10~point type, with a vertical spacing of 11~points.
% The word \textsc{Abstract} must be centered, in small caps, and in point size 12. Two
% line spaces precede the abstract. 
\end{abstract}

\section{Introduction}

Distillation is a widely used approach for compressing large models into smaller efficient counterparts while preserving most of their predictive power~\cite{hinton2015distillingknowledgeneuralnetwork, romero2015fitnetshintsdeepnets, zagoruyko2017payingattentionattentionimproving, Yim2017CVPR, tung2019similaritypreservingknowledgedistillation, heo2019comprehensiveoverhaulfeaturedistillation}. As model sizes continue to grow, both training and inference become increasingly expensive and computationally demanding. Larger models also suffer from slower inference times, which is problematic in domains requiring large-scale forward passes. For instance, in-silico cell perturbation studies often involve evaluating thousands or potentially millions of computational experiments~\cite{lotfollahi2019scgen, lopez2018deep, cui2024scgpt}. Compact yet capable models are therefore essential for feasibility and cost-effectiveness. Such models can also serve as efficient proxies for rapid verification or filtering before resorting to billion-parameter architectures.

In this paper, we use distillation to train a student model called HelixNano-mRNA (Hybrid Mamba-Attention model~\cite{wood2025helixmrnahybridfoundationmodel} with $\sim$5M parameters) using representations of mRNA sequences from a teacher model, which we chose to be Evo2-1B~\cite{Brixi2025.02.18.638918}. Specifically, we align intermediate embeddings from Evo2 with two student hidden layers to transfer structural and contextual information. While the teacher model is trained on both DNA and RNA sequences, we distill knowledge only on mRNA inputs to obtain a student model specialized for mRNA representation learning. Direct logit distillation~\cite{hinton2015distillingknowledgeneuralnetwork} was found to be unstable, with the KL divergence oscillating heavily during training~\cite{yuan2021revisitingknowledgedistillationlabel}. This may be due to our work involving biological data as opposed to text~\cite{jiao2020tinybertdistillingbertnatural}. Embedding distillation produced consistent improvements yielding state-of-the-art performance among models of comparable size and out-performing a host of much larger models~\cite{sun2019patientknowledgedistillationbert, jiao2020tinybertdistillingbertnatural}.

Little work exists in the post-training landscape for genomic models, and even less so for distillation of these models~\cite{madani2020progenlanguagemodelingprotein, geffen2022distilprotbert}. In this work we show the following contributions:
\begin{itemize}
    \item Embedding-based distillation, at least for mRNA sequences, is more stable than logit-based distillation for genomic models.
    \item 200-Fold reduction in size while achieving state-of-the-art performance for its size on \texttt{mRNA-bench}~\cite{shi_dalal_fradkin_2025_mrnabench}.
    \item Using intermediate latent representations is an effective approach for biological model distillation.
\end{itemize}

After the distillation, we name the resulting student model HelixNano-mRNA. It can be easily extended with linear layers and a task-specific loss for downstream mRNA sequence generation, classification, and regression. We release the distilled model weights at \url{https://huggingface.co/helical-ai/HelixNano-mRNA}.

\section{Methods}
\label{gen_inst}

\subsection{Distillation via Intermediary Latent Embeddings}
In general, the distillation loss function can be expressed as a combination of $\mathcal{L}_{\text{total}} = {\mathcal{L}_{\mathrm{CE}}} + {\mathcal{L}_{\mathrm{KL}} + \mathcal{L}_{\mathrm{ED}}}$, where $\mathcal{L}_{\mathrm{KL}} = \mathrm{KL}\big(p_t(x) \,\|\, p_s(x)\big)$ is the KL divergence between the output probability distributions of the teacher $p_t(x)$ and student $p_s(x)$~\cite{hinton2015distillingknowledgeneuralnetwork, gou2021knowledge}.
$\mathcal{L}_{\mathrm{ED}}$ is the loss between teacher $E_t(x)$ and student embeddings $E_s(x)$. In our case specifically, we define $\mathcal{L}_{\mathrm{ED}}$ as a combination of both the mean-square loss $\mathcal{L}_{\mathrm{mse}} = \lVert ||E_t(x) - E_s(x)|| \rVert_2^2$ and cosine  loss $\mathcal{L}_{\mathrm{cos}} = 1 - E_t(x) \cdot E_s(x)$. Empirically in our experiments, removing the KL and cross-entropy terms performed best which is why in the following we only use the embedding matching loss.  

Our adapted distillation loss is given by $\mathcal{L}_{\text{train}} = \lambda_{\mathrm{cos}} \,\mathcal{L}_{\mathrm{cos}} + \lambda_{\mathrm{mse}} \,\mathcal{L}_{\mathrm{mse}} + \lambda_{\mathrm{wd}} \,\lVert \theta \rVert_2^2$, where $\mathcal{L}_{\mathrm{cos}}$ is the cosine embedding loss between up-projected student embeddings and teacher embeddings and $\theta$ denotes all trainable student parameters. We set $\lambda_{\mathrm{cos}} \gg \lambda_{\mathrm{mse}}$ to ensure directional alignment dominates while the Euclidean term prevents exploding norms. Alongside optimizer weight decay this led to better training stability and smaller absolute embedding values.

Focusing on mRNA sequences, we first tokenize mRNA sequences similarly to the original Evo2-1B model (per nucleotide) and feed them to both models, freezing the teachers weights. 
% In its most general form, the student would learn using the standard task loss but also from additional alignment signals from the teacher’s embeddings and logits. 

\begin{figure}[t]
  \centering
  \includegraphics[width=0.5\linewidth]{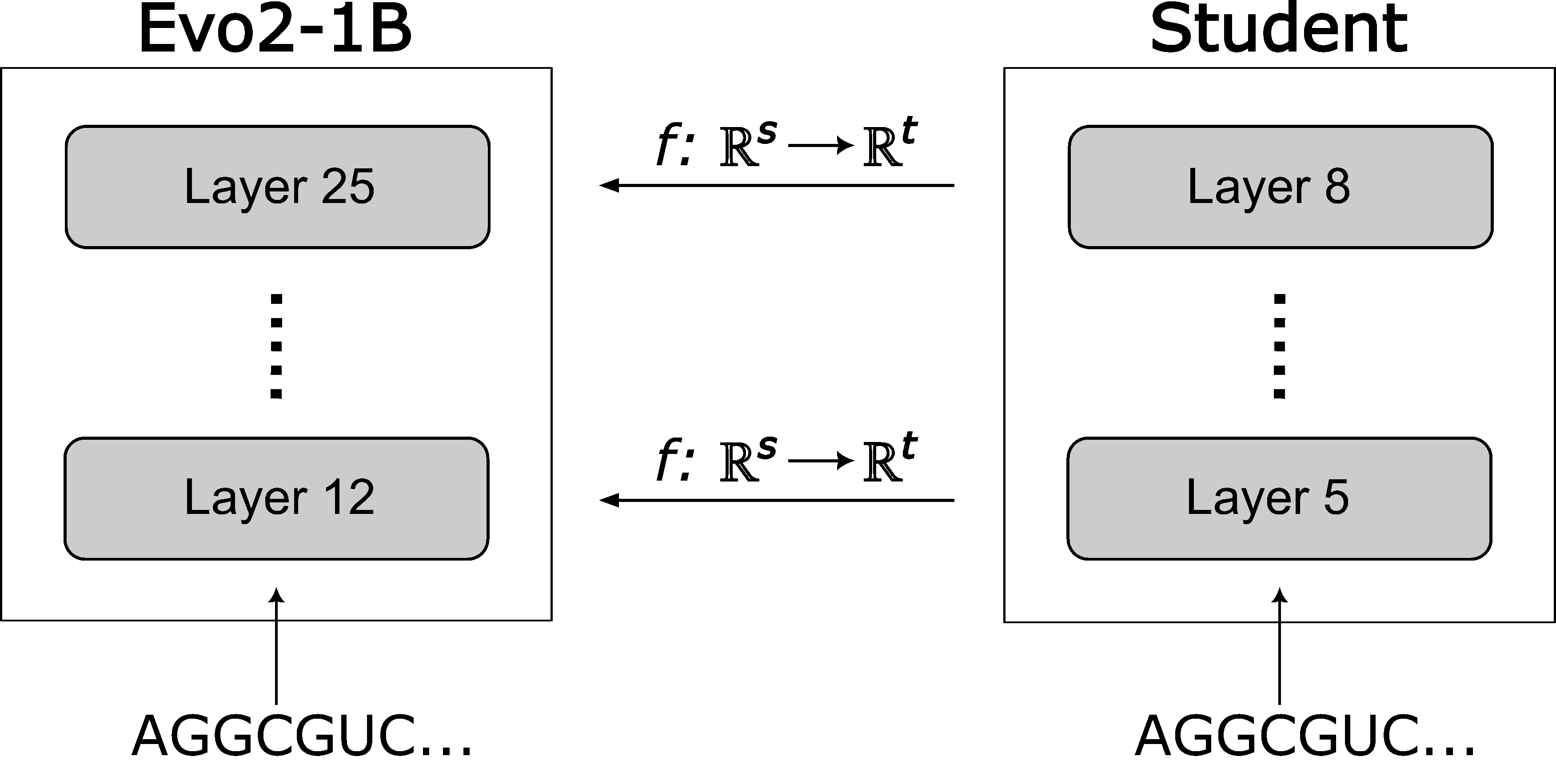}
  \caption{The student model is aligned with two hidden layers (5th and 8th) using projections from layers of Evo2-1B (12th and 25th) .}
  \label{fig:main_sketch}
\end{figure}

To match the different sizes between the teacher and student model we use linear projection layers to align the hidden dimensions of latent teacher and student embeddings (see Figure~\ref{fig:main_sketch}, $\mathbb{R}^{256} \rightarrow \mathbb{R}^{1942}$)~\cite{tian2022contrastiverepresentationdistillation, zbontar2021barlowtwinsselfsupervisedlearning, bardes2022vicregvarianceinvariancecovarianceregularizationselfsupervised, chen2023improvedfeaturedistillationprojector, miles2024understandingroleprojectorknowledge}. Down-projection of the teacher embeddings can lead to zero loss solutions as both the model and projection layers collapse. We did not choose layers too early in Evo2 as they may not encode as much latent information and found two layer matching to work better than one layer~\cite{sun2019patientknowledgedistillationbert}. Prior work on Genomics Foundation Models has also shown that the best embeddings are not typically from final layers~\cite{DallaTorre2025NucleotideTransformer, Brixi2025.02.18.638918}. For student embeddings we concatenate the output from both layers and the mean over the context length is taken for a one-dimensional final embedding.

As we match final layers of both models, the student may have learnt sufficient representations from Evo2 for mapping back to the token space (e.g., using an additional linear layer). Due to computational limits, we were not able to evaluate how all combinations of layer pairs affect evaluation metrics. Nevertheless, we achieve state-of-the-art performance on a current benchmark using reasonable chosen blocks~\cite{shi_dalal_fradkin_2025_mrnabench}. We found that taking the embeddings post-norm worked better due to normalisation. We do not expect our model to learn all Evo2 latent representations due to the dimension mismatch but rather learn to encode a number of essential directions.

\subsection{Distillation Details}
We train with the following hyperparameters: batch size 32, sequence lengths of 2048 tokens, AdamW with a weight decay of $1\times 10^{-2}$, learning rate of $2\times 10^{-4}$ with linear warmup over the first 2000 steps, $\lambda_{\mathrm{cos}} = 1.0$, $\lambda_{\mathrm{mse}} = 0.1$, dropout $10\%$, and gradient norm clipping with a maximum of $1.0$. We train in mixed precision (bfloat16) on 4x A100 GPUs. Total training time is 24 hours.  

\subsection{Dataset}

Training mRNA datasets can be found on the NCBI ftp server: \url{https://ftp.ncbi.nih.gov/
refseq/release/}. We used all the files ending in '.rna.gbff.gz' and subsampled 27 million sequences out of the possible 56 million due to budgetary constraints.
The dataset consisted of 43.6\% other vertebrates (\url{https://ftp.ncbi.nih.gov/refseq/release/vertebrate_other/}), 28.3\% mammals (\url{https://ftp.ncbi.nih.gov/refseq/release/vertebrate_mammalian/}), 26.4\% invertebrates (\url{https://ftp.ncbi.nih.gov/refseq/release/invertebrate/}) and 1.6\% viruses (\url{https://ftp.ncbi.nih.gov/refseq/release/viral/}).

\subsection{Evaluation and Metrics}
During training and evaluation we monitor the below metrics (alongside total loss). Note we use an up-arrow ($\uparrow$) for student embeddings \textbf{after} the linear projection layer.

To assess the effect of the projection layer, we applied the above Centred Kernel Alignment (CKA) metric, which quantifies the structural similarity between two sets of representations (independent of embedding dimensionality~\cite{kornblith2019similarityneuralnetworkrepresentations}.

In total we use the following metrics: \textbf{Projection cosine loss and MSE} for each aligned layer $\mathcal{L}_{\cos} = 1 - E_s^{\uparrow}(x) \cdot E_t(x)$ and $\text{MSE} = \frac{1}{n}\sum_{i=1}^{n} \left(E_s^{\uparrow}(x_i) -E_t(x_i)\right)^2$; \textbf{Student embedding variance} to detect collapse: $\text{Var}(E_s) = \frac{1}{d} \sum_{j=1}^{d} \text{Var}\!\left(E_{s,j}\right)$; \textbf{Linear CKA} between teacher embeddings and (up-projected) student embeddings $ \text{CKA}(E_t,\, E_s^{(\uparrow)}) = \frac{\left\|E_t^\top E_s^{(\uparrow)}\right\|_F^2}{\left\|E_t^\top E_t\right\|_F \;\left\| (E_s^{(\uparrow)})^\top E_s^{(\uparrow)} \right\|_F}$.

After training, we benchmark the model on \texttt{mRNA-bench}~\cite{shi_dalal_fradkin_2025_mrnabench}, which involves : mRNA Half-Life (\textbf{HL}) (measures transcript stability via half-life), mRNA Subcellular Localization (\textbf{mRNA-Loc-SR}) (predicts cellular compartment from short-read data; long-read data unavailable \footnote{\url{https://github.com/morrislab/mRNABench/issues/23}}), Paired mRNA Half-Life and Mean Ribosome Load (\textbf{MRL-HL-Pair}) (jointly models stability and translation efficiency), GO Term Classification (\textbf{GO}) (predicts gene function across molecular, biological, and cellular categories), Protein Localization (\textbf{Prot-Loc}) (predicts protein subcellular localization)
Massively Parallel Translation Assay – Mean Ribosome Load (\textbf{MRL-MPRA}) (predicts translation efficiency from synthetic 5'UTRs), eCLIP Binding (\textbf{eCLIP}) (predicts RNA-binding protein interactions), and Variant Effect Prediction (\textbf{VEP}) (predicts pathogenic single-nucleotide variants in mRNA).

\section{Results}
\label{headings}

We train until convergence with training and validation curves shown on Figure~\ref{fig:loss_curves}. Both losses decrease sharply during the initial phase and then reduce much slower. Similar behaviour is reflected in other metrics (see Appendix). During training gradient norms and variances exploded using only a cosine loss, which is why a small MSE loss was added between student and teacher embeddings (see Figures~\ref{fig:norms}--\ref{fig:mse_losses}). We computed CKA both before and after projection to evaluate whether the linear layer contributes substantially. Figure~\ref{fig:cka_plots} shows the CKA values pre- and post-projection, revealing no significant impact from the linear layer.

\begin{figure}[h!]
    \centering
    \begin{subfigure}[c]{0.48\textwidth}
        \centering
        \includegraphics[width=\linewidth]{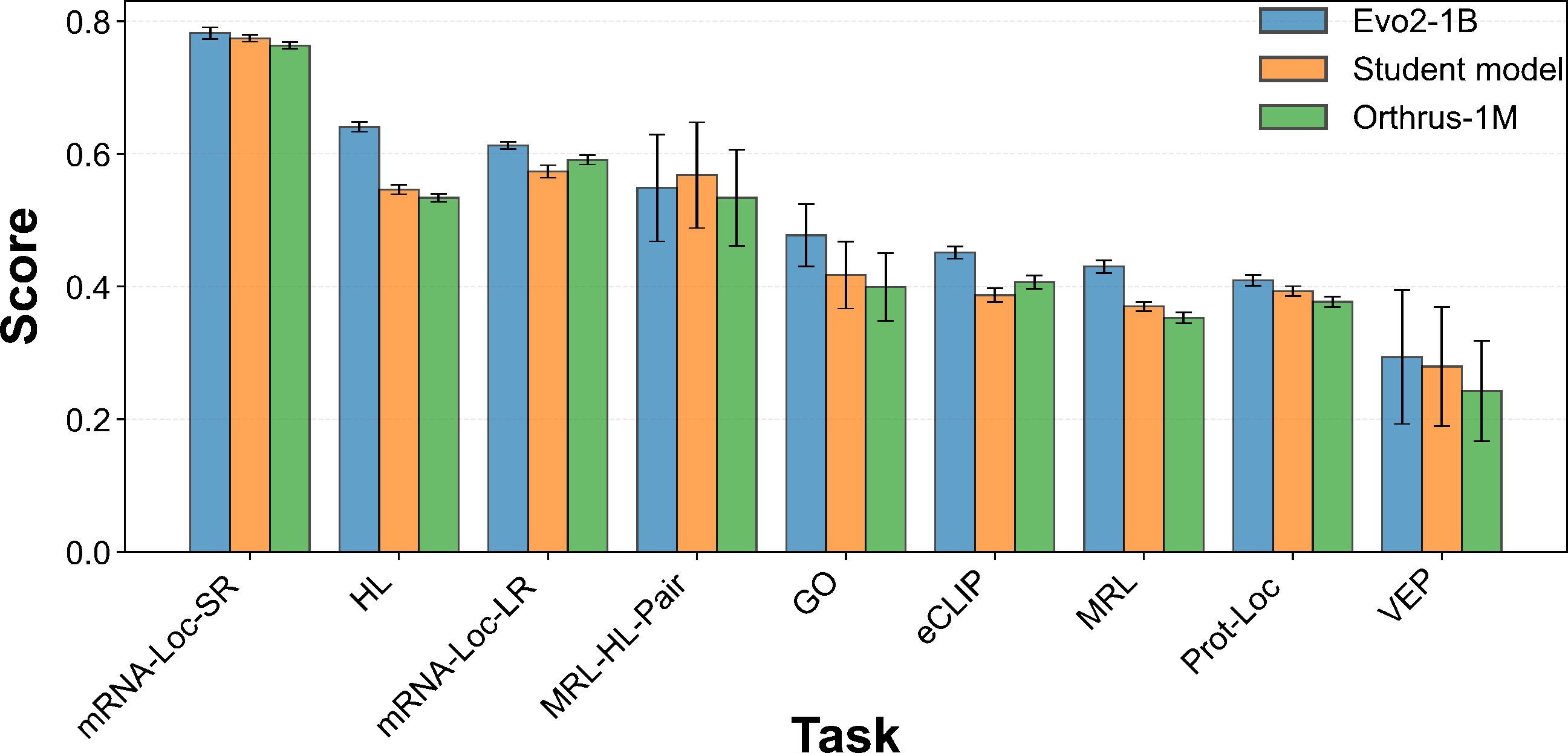}
        \caption{Comparing models by task.}
        \label{fig:by_task}
    \end{subfigure}
    \hfill
    \begin{subfigure}[c]{0.48\textwidth}
        \centering
        \includegraphics[width=\linewidth]{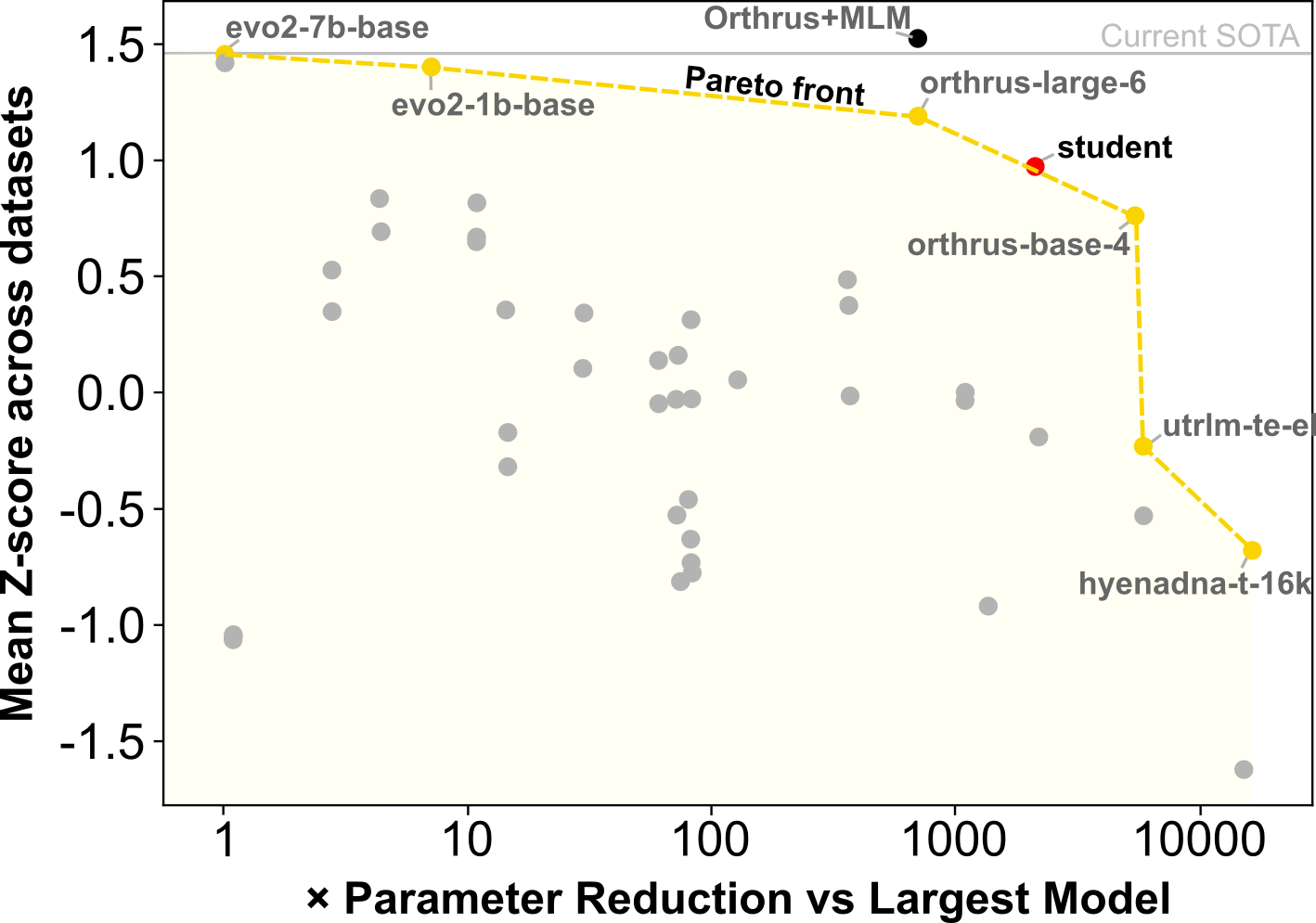}
        \caption{Overall model performance by size. Our model is denoted by red dot. Figure adapted and re-created from~\cite{shi_dalal_fradkin_2025_mrnabench}.}
        \label{fig:benchmark_results}
    \end{subfigure}
    \caption{Benchmark results.}
    \label{fig:benchmarks}
\end{figure}

We benchmarked Evo2-1B, the student model - HelixNano-mRNA, and orthrus-base-4 (Orthrus-1M)~\cite{fradkin2024orthrus} on \texttt{mRNA-bench}. Score by task are shown in Figure~\ref{fig:by_task} while overall score by model size is shown on Figure~\ref{fig:benchmark_results}. The distilled model achieves leading performance across nearly all tasks for its size \footnote{mRNA-Loc-LR omitted from aggregated results due to unavailability of the processed version used in~\cite{shi_dalal_fradkin_2025_mrnabench} at time of writing; see \url{https://github.com/morrislab/mRNABench/issues/23}}.

% \begin{figure}[h!]
%     \centering
%         \includegraphics[width=0.7\linewidth]{figures/bench_by_task.png}
%         \caption{Comparing models by task.}
%         \label{fig:by_task}
% \end{figure}

% \begin{figure}[t]
%     \centering
%     % \vspace{-10pt} % optional: pulls figure up a bit
%     \includegraphics[width=0.6\linewidth]{figures/benchmark_all_models.png}
%     \caption{Overall model performance by size. Our model is denoted by red dot. Figure adapted and re-created from~\cite{shi_dalal_fradkin_2025_mrnabench}.}
%     \label{fig:benchmark_results}
%     % \vspace{-10pt} % optional: reduces space after
% \end{figure}

% for camera ready comment the above figure and uncomment the below so its properly wrapped. 
% \begin{wrapfigure}{r}{0.72\linewidth}
%     \centering
%     % \vspace{-10pt} % optional: pulls figure up a bit
%     \includegraphics[width=\linewidth]{figures/benchmark_all_models.png}
%     \caption{Overall model performance by size. Figure adapted and re-created from \cite{shi_dalal_fradkin_2025_mrnabench}.}
%     \label{fig:benchmark_results}
%     \vspace{-10pt} % optional: reduces space after
% \end{wrapfigure}

Distillation was attempted using sequence logits and KL divergences but this did not perform as well. The entropy for the logit distribution was very noisy, including large spikes and erratic behaviour, making model learning difficult particularly given its smaller size (see Figure~\ref{fig:entropy_noise} and Figure~\ref{fig:summary_ent}). We briefly tested the model by matching a single hidden layer with Evo2 but found better results with two layer matching. This is not an extensive analysis and more work should be done to verify this behaviour~\cite{yu2025revisitingintermediatelayermatchingknowledge}. 

For the matched layers we implemented PCA to the Evo2 embeddings giving us the number of dimensions needed to account for $>90\%$ of the variance. This was a surprisingly small for post-norm layers (6 dimensions) but much higher for block 12 (431, see Table~\ref{tab:pca}) which was used for downstream biological tasks by the Evo2 authors~\cite{Brixi2025.02.18.638918}. The norm layers are most likely collapsing the embedding space, explaining why the student model was able to achieve lower cosine losses as compared to non-norm layers. A UMAP of embeddings by phylum and model is shown in the Appendix (see Figure~\ref{fig:umap}).

\section{Future Work}

This work presents HelixNano-mRNA and aims to set up foundations for distillation and other post-training work in the biological foundational model domain. Future work can focus on both further interpreting noisy KL divergences in distillation of biological sequences alongside effects of matching various layers. Extending the distilled embeddings to downstream tasks, such as sequence generation and variant effect prediction will further evaluate predictive performance.

\FloatBarrier
\bibliography{gen2_iclr2026_workshop}

@misc{hinton2015distillingknowledgeneuralnetwork,
      title={Distilling the Knowledge in a Neural Network}, 
      author={Geoffrey Hinton and Oriol Vinyals and Jeff Dean},
      year={2015},
      eprint={1503.02531},
      archivePrefix={arXiv},
      primaryClass={stat.ML},
      url={https://arxiv.org/abs/1503.02531}, 
}

@misc{romero2015fitnetshintsdeepnets,
      title={FitNets: Hints for Thin Deep Nets}, 
      author={Adriana Romero and Nicolas Ballas and Samira Ebrahimi Kahou and Antoine Chassang and Carlo Gatta and Yoshua Bengio},
      year={2015},
      eprint={1412.6550},
      archivePrefix={arXiv},
      primaryClass={cs.LG},
      url={https://arxiv.org/abs/1412.6550}, 
}

@misc{zagoruyko2017payingattentionattentionimproving,
      title={Paying More Attention to Attention: Improving the Performance of Convolutional Neural Networks via Attention Transfer}, 
      author={Sergey Zagoruyko and Nikos Komodakis},
      year={2017},
      eprint={1612.03928},
      archivePrefix={arXiv},
      primaryClass={cs.CV},
      url={https://arxiv.org/abs/1612.03928}, 
}

@InProceedings{Yim2017CVPR,
author = {Yim, Junho and Joo, Donggyu and Bae, Jihoon and Kim, Junmo},
title = {A Gift From Knowledge Distillation: Fast Optimization, Network Minimization and Transfer Learning},
booktitle = {Proceedings of the IEEE Conference on Computer Vision and Pattern Recognition (CVPR)},
month = {July},
year = {2017}
}

@misc{tung2019similaritypreservingknowledgedistillation,
      title={Similarity-Preserving Knowledge Distillation}, 
      author={Frederick Tung and Greg Mori},
      year={2019},
      eprint={1907.09682},
      archivePrefix={arXiv},
      primaryClass={cs.CV},
      url={https://arxiv.org/abs/1907.09682}, 
}

@misc{heo2019comprehensiveoverhaulfeaturedistillation,
      title={A Comprehensive Overhaul of Feature Distillation}, 
      author={Byeongho Heo and Jeesoo Kim and Sangdoo Yun and Hyojin Park and Nojun Kwak and Jin Young Choi},
      year={2019},
      eprint={1904.01866},
      archivePrefix={arXiv},
      primaryClass={cs.CV},
      url={https://arxiv.org/abs/1904.01866}, 
}

@misc{yuan2021revisitingknowledgedistillationlabel,
      title={Revisiting Knowledge Distillation via Label Smoothing Regularization}, 
      author={Li Yuan and Francis E. H. Tay and Guilin Li and Tao Wang and Jiashi Feng},
      year={2021},
      eprint={1909.11723},
      archivePrefix={arXiv},
      primaryClass={cs.CV},
      url={https://arxiv.org/abs/1909.11723}, 
}

@article{lopez2018deep,
  title={Deep generative modeling for single-cell transcriptomics},
  author={Lopez, Romain and Regier, Jeffrey and Cole, Michael B and Jordan, Michael I and Yosef, Nir},
  journal={Nature Methods},
  volume={15},
  number={12},
  pages={1053--1058},
  year={2018},
  doi={10.1038/s41592-018-0229-2}
}

@article{lotfollahi2019scgen,
  title={scGen predicts single-cell perturbation responses},
  author={Lotfollahi, Mohammad and Wolf, Fabian A and Theis, Fabian J},
  journal={Nature Methods},
  volume={16},
  number={8},
  pages={715--721},
  year={2019},
  doi={10.1038/s41592-019-0494-8}
}

@article{cui2024scgpt,
  title={scGPT: Towards building a foundation model for single-cell multi-omics using generative AI},
  author={Cui, Haotian and Wang, Chloe and Maan, Hassaan and Pang, Kuan and Luo, Fengning and Wang, Bo},
  journal={Nature Methods},
  year={2024},
  doi={10.1038/s41592-024-02201-0}
}

@misc{wood2025helixmrnahybridfoundationmodel,
      title={Helix-mRNA: A Hybrid Foundation Model For Full Sequence mRNA Therapeutics}, 
      author={Matthew Wood and Mathieu Klop and Maxime Allard},
      year={2025},
      eprint={2502.13785},
      archivePrefix={arXiv},
      primaryClass={q-bio.GN},
      url={https://arxiv.org/abs/2502.13785}, 
}

@article{DallaTorre2025NucleotideTransformer,
  title     = {Nucleotide Transformer: building and evaluating robust foundation models for human genomics},
  author    = {Dalla-Torre, Hugo and Gonzalez, Liam and Mendoza-Revilla, Javier and Lopez Carranza, Nicolas and Grzywaczewski, Adam Henryk and Oteri, Francesco and Dallago, Christian and Trop, Evan and de Almeida, Bernardo P. and Sirelkhatim, Hassan and Richard, Guillaume and Skwark, Marcin and Beguir, Karim and Lopez, Marie and Pierrot, Thomas},
  journal   = {Nature Methods},
  year      = {2025},
  volume    = {22},
  number    = {2},
  pages     = {287--297},
  doi       = {10.1038/s41592-024-02523-z},
  pmid      = {39609566},
  pmcid     = {PMC11810778},
  note      = {Epub 2024 Nov 28}
}

@article{Brixi2025.02.18.638918,
	author = {Brixi, Garyk and Durrant, Matthew G and Ku, Jerome and Poli, Michael and Brockman, Greg and Chang, Daniel and Gonzalez, Gabriel A and King, Samuel H and Li, David B and Merchant, Aditi T and Naghipourfar, Mohsen and Nguyen, Eric and Ricci-Tam, Chiara and Romero, David W and Sun, Gwanggyu and Taghibakshi, Ali and Vorontsov, Anton and Yang, Brandon and Deng, Myra and Gorton, Liv and Nguyen, Nam and Wang, Nicholas K and Adams, Etowah and Baccus, Stephen A and Dillmann, Steven and Ermon, Stefano and Guo, Daniel and Ilango, Rajesh and Janik, Ken and Lu, Amy X and Mehta, Reshma and Mofrad, Mohammad R.K. and Ng, Madelena Y and Pannu, Jaspreet and Re, Christopher and Schmok, Jonathan C and St. John, John and Sullivan, Jeremy and Zhu, Kevin and Zynda, Greg and Balsam, Daniel and Collison, Patrick and Costa, Anthony B. and Hernandez-Boussard, Tina and Ho, Eric and Liu, Ming-Yu and McGrath, Tom and Powell, Kimberly and Burke, Dave P. and Goodarzi, Hani and Hsu, Patrick D and Hie, Brian},
	title = {Genome modeling and design across all domains of life with Evo 2},
	elocation-id = {2025.02.18.638918},
	year = {2025},
	doi = {10.1101/2025.02.18.638918},
	publisher = {Cold Spring Harbor Laboratory},
	URL = {https://www.biorxiv.org/content/early/2025/02/21/2025.02.18.638918},
	eprint = {https://www.biorxiv.org/content/early/2025/02/21/2025.02.18.638918.full.pdf},
	journal = {bioRxiv}
}

@misc{tian2022contrastiverepresentationdistillation,
      title={Contrastive Representation Distillation}, 
      author={Yonglong Tian and Dilip Krishnan and Phillip Isola},
      year={2022},
      eprint={1910.10699},
      archivePrefix={arXiv},
      primaryClass={cs.LG},
      url={https://arxiv.org/abs/1910.10699}, 
}

@article{shi_dalal_fradkin_2025_mrnabench,
    author = {Shi, Ruian and Dalal, Taykhoom and Fradkin, Philip and Koyyalagunta, Divya and Chhabria, Simran and Jung, Andrew and Tam, Cyrus and Ceyhan, Defne and Lin, Jessica and Laverty, Kaitlin U. and Baali, Ilyes and Wang, Bo and Morris, Quaid},
    title = {mRNABench: A curated benchmark for mature mRNA property and function prediction},
    elocation-id = {2025.07.05.662870},
    year = {2025},
    doi = {10.1101/2025.07.05.662870},
    publisher = {Cold Spring Harbor Laboratory},
    URL = {https://www.biorxiv.org/content/early/2025/07/08/2025.07.05.662870},
    eprint = {https://www.biorxiv.org/content/early/2025/07/08/2025.07.05.662870.full.pdf},
    journal = {bioRxiv}
}

@misc{chen2023improvedfeaturedistillationprojector,
      title={Improved Feature Distillation via Projector Ensemble}, 
      author={Yudong Chen and Sen Wang and Jiajun Liu and Xuwei Xu and Frank de Hoog and Zi Huang},
      year={2023},
      eprint={2210.15274},
      archivePrefix={arXiv},
      primaryClass={cs.CV},
      url={https://arxiv.org/abs/2210.15274}, 
}

@misc{kornblith2019similarityneuralnetworkrepresentations,
      title={Similarity of Neural Network Representations Revisited}, 
      author={Simon Kornblith and Mohammad Norouzi and Honglak Lee and Geoffrey Hinton},
      year={2019},
      eprint={1905.00414},
      archivePrefix={arXiv},
      primaryClass={cs.LG},
      url={https://arxiv.org/abs/1905.00414}, 
}

@article{fradkin2024orthrus,
  title={Orthrus: Towards Evolutionary and Functional RNA Foundation Models},
  author={Fradkin, Philip and Shi, Ruian and Isaev, Keren and Frey, Brendan J. and Morris, Quaid and Lee, Leo J. and Wang, Bo},
  journal={bioRxiv},
  year={2024},
  doi={10.1101/2024.10.10.617658},
}

@misc{sun2019patientknowledgedistillationbert,
      title={Patient Knowledge Distillation for BERT Model Compression}, 
      author={Siqi Sun and Yu Cheng and Zhe Gan and Jingjing Liu},
      year={2019},
      eprint={1908.09355},
      archivePrefix={arXiv},
      primaryClass={cs.CL},
      url={https://arxiv.org/abs/1908.09355}, 
}

@misc{jiao2020tinybertdistillingbertnatural,
      title={TinyBERT: Distilling BERT for Natural Language Understanding}, 
      author={Xiaoqi Jiao and Yichun Yin and Lifeng Shang and Xin Jiang and Xiao Chen and Linlin Li and Fang Wang and Qun Liu},
      year={2020},
      eprint={1909.10351},
      archivePrefix={arXiv},
      primaryClass={cs.CL},
      url={https://arxiv.org/abs/1909.10351}, 
}

@misc{yu2025revisitingintermediatelayermatchingknowledge,
      title={Revisiting Intermediate-Layer Matching in Knowledge Distillation: Layer-Selection Strategy Doesn't Matter (Much)}, 
      author={Zony Yu and Yuqiao Wen and Lili Mou},
      year={2025},
      eprint={2502.04499},
      archivePrefix={arXiv},
      primaryClass={cs.LG},
      url={https://arxiv.org/abs/2502.04499}, 
}

@misc{zbontar2021barlowtwinsselfsupervisedlearning,
      title={Barlow Twins: Self-Supervised Learning via Redundancy Reduction}, 
      author={Jure Zbontar and Li Jing and Ishan Misra and Yann LeCun and Stéphane Deny},
      year={2021},
      eprint={2103.03230},
      archivePrefix={arXiv},
      primaryClass={cs.CV},
      url={https://arxiv.org/abs/2103.03230}, 
}

@misc{bardes2022vicregvarianceinvariancecovarianceregularizationselfsupervised,
      title={VICReg: Variance-Invariance-Covariance Regularization for Self-Supervised Learning}, 
      author={Adrien Bardes and Jean Ponce and Yann LeCun},
      year={2022},
      eprint={2105.04906},
      archivePrefix={arXiv},
      primaryClass={cs.CV},
      url={https://arxiv.org/abs/2105.04906}, 
}

@misc{miles2024understandingroleprojectorknowledge,
      title={Understanding the Role of the Projector in Knowledge Distillation}, 
      author={Roy Miles and Krystian Mikolajczyk},
      year={2024},
      eprint={2303.11098},
      archivePrefix={arXiv},
      primaryClass={cs.CV},
      url={https://arxiv.org/abs/2303.11098}, 
}

@misc{madani2020progenlanguagemodelingprotein,
      title={ProGen: Language Modeling for Protein Generation}, 
      author={Ali Madani and Bryan McCann and Nikhil Naik and Nitish Shirish Keskar and Namrata Anand and Raphael R. Eguchi and Po-Ssu Huang and Richard Socher},
      year={2020},
      eprint={2004.03497},
      archivePrefix={arXiv},
      primaryClass={q-bio.BM},
      url={https://arxiv.org/abs/2004.03497}, 
}

@misc{geffen2022distilprotbert,
  title  = {DistilProtBert: a distilled protein language model used to distinguish between real proteins and their randomly shuffled counterparts},
  author = {Yaron Geffen and Yanay Ofran and Ron Unger},
  year   = {2022},
  doi    = {10.1093/bioinformatics/btac474}
}

@article{gou2021knowledge,
  title={Knowledge distillation: A survey},
  author={Gou, Jianping and Yu, Baosheng and Maybank, Stephen J and Tao, Dacheng},
  journal={International journal of computer vision},
  volume={129},
  number={6},
  pages={1789--1819},
  year={2021},
  publisher={Springer}
}
\bibliographystyle{mlgenx_conference}

\newpage
\appendix
\makeatletter
\setlength{\@fptop}{0pt}
\makeatother
\FloatBarrier

\section{Appendix}
\setcounter{figure}{0}
\renewcommand{\thefigure}{A\arabic{figure}}

\begin{figure}[htbp]
    \centering
    \begin{subfigure}{0.49\linewidth}
        \centering
        \includegraphics[width=\linewidth]{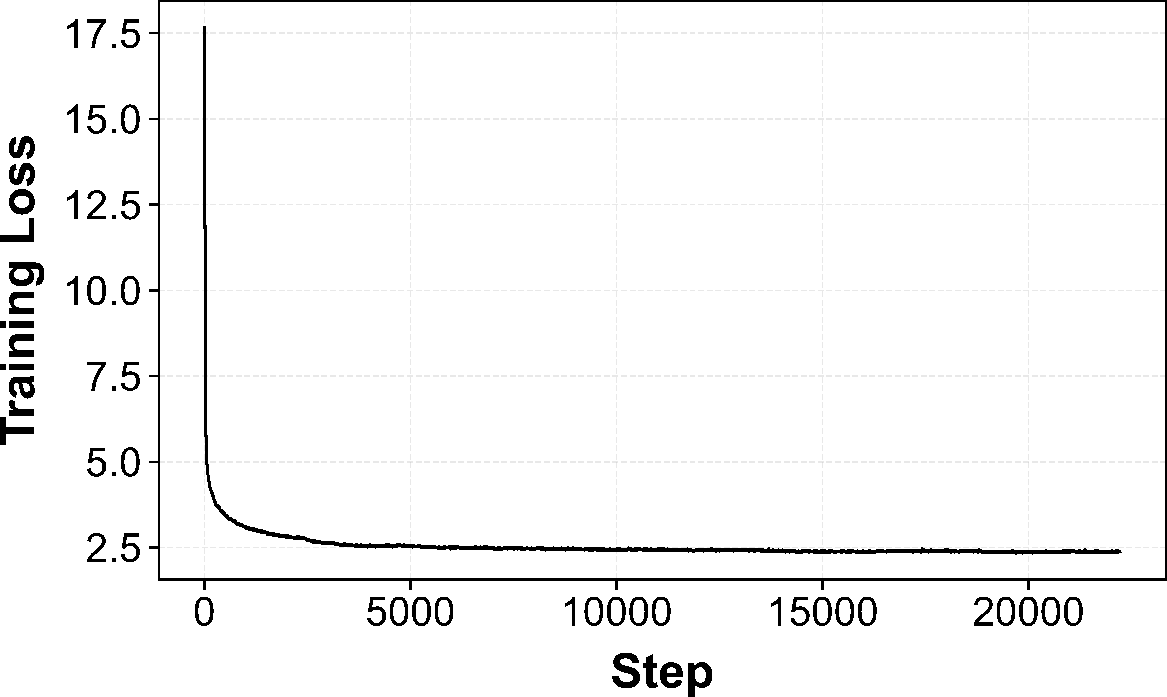}
    \end{subfigure}
    \hfill
    \begin{subfigure}{0.49\linewidth}
        \centering
        \includegraphics[width=\linewidth]{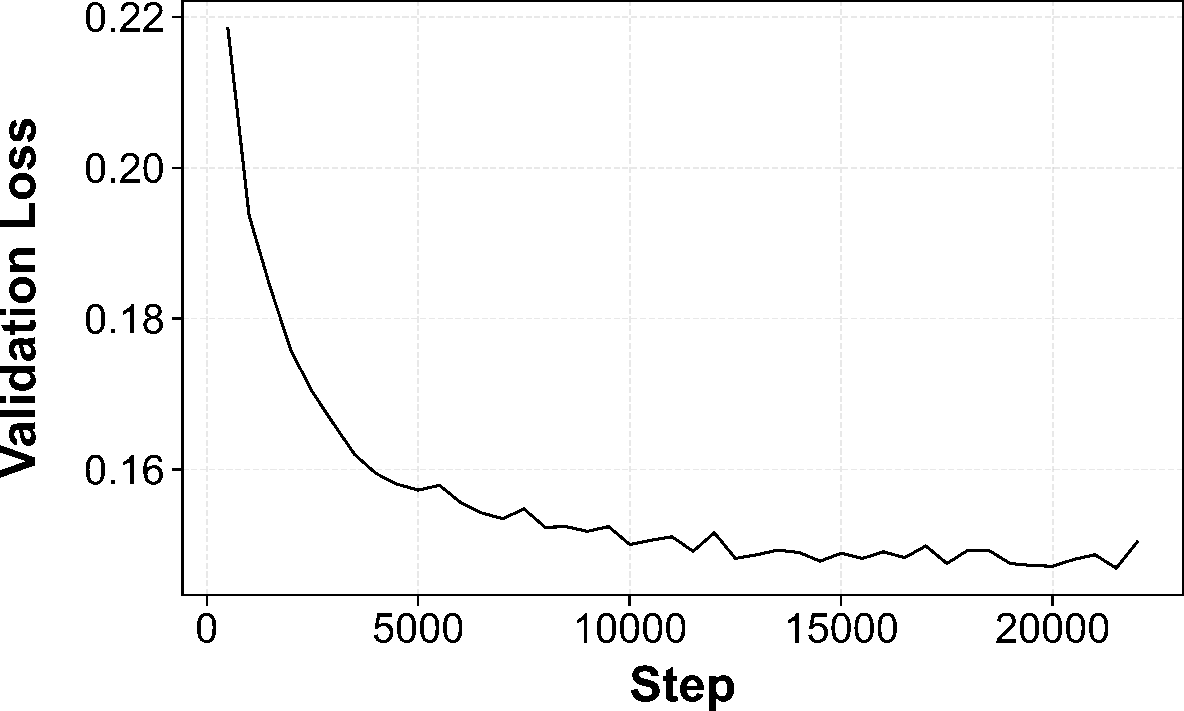}
    \end{subfigure}
    \caption{Losses for the train and validation set}
    \label{fig:loss_curves}
\end{figure}

\begin{figure}[htbp]
    \centering
    \begin{subfigure}[t]{0.49\linewidth}
        \centering
        \includegraphics[width=\linewidth]{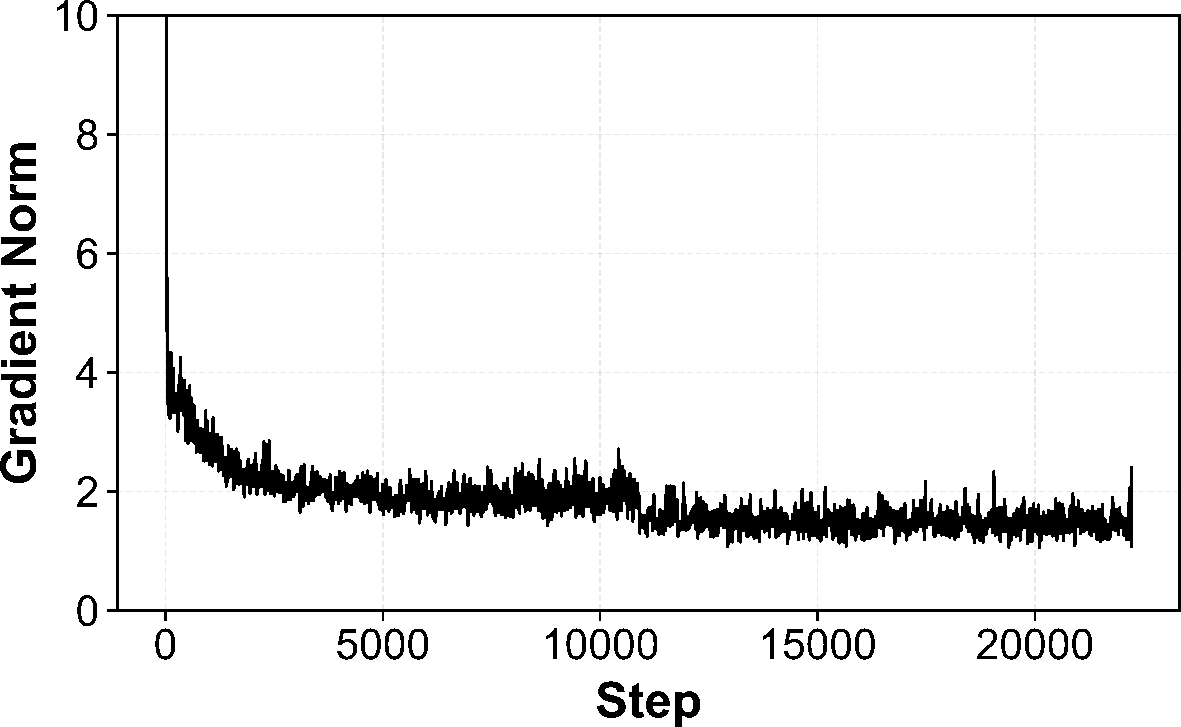}
    \end{subfigure}
    \hfill
    \begin{subfigure}[t]{0.49\linewidth}
        \centering
        \includegraphics[width=\linewidth]{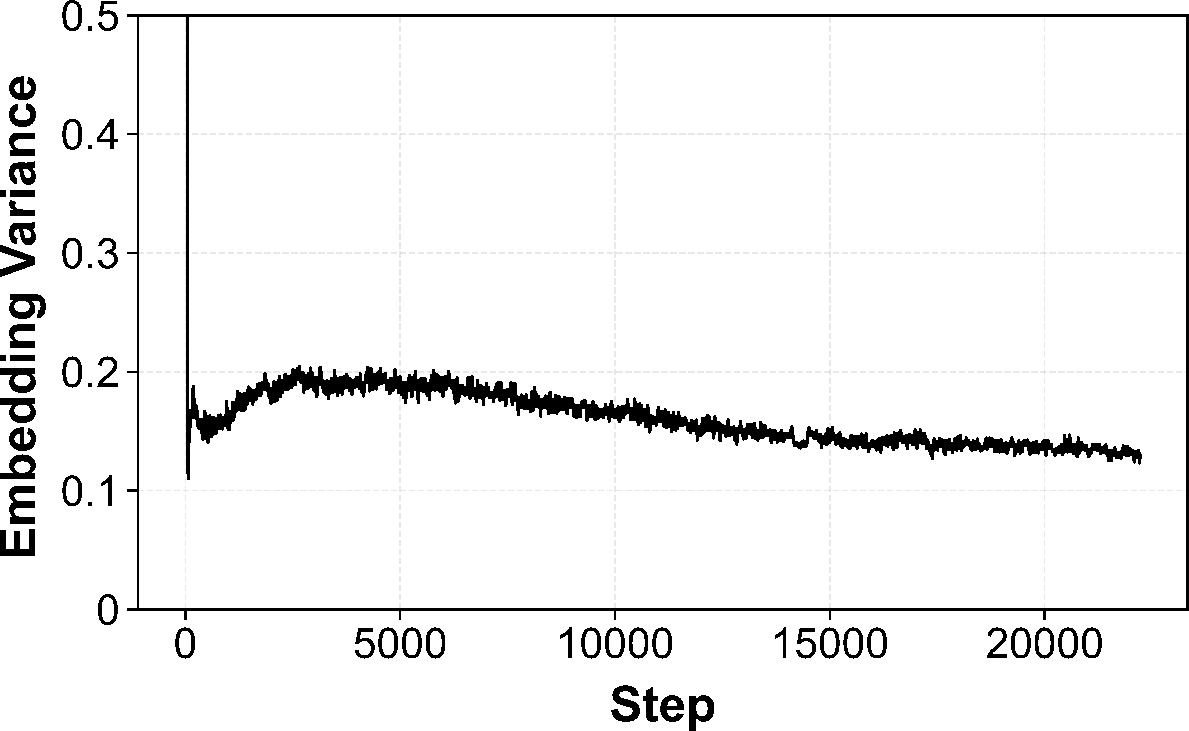}
    \end{subfigure}
    \caption{Student model gradient norms and embedding variances during training}
    \label{fig:norms}
\end{figure}

\begin{figure}[htbp]
    \centering
    \begin{subfigure}[t]{0.49\linewidth}
        \centering
        \includegraphics[width=\linewidth]{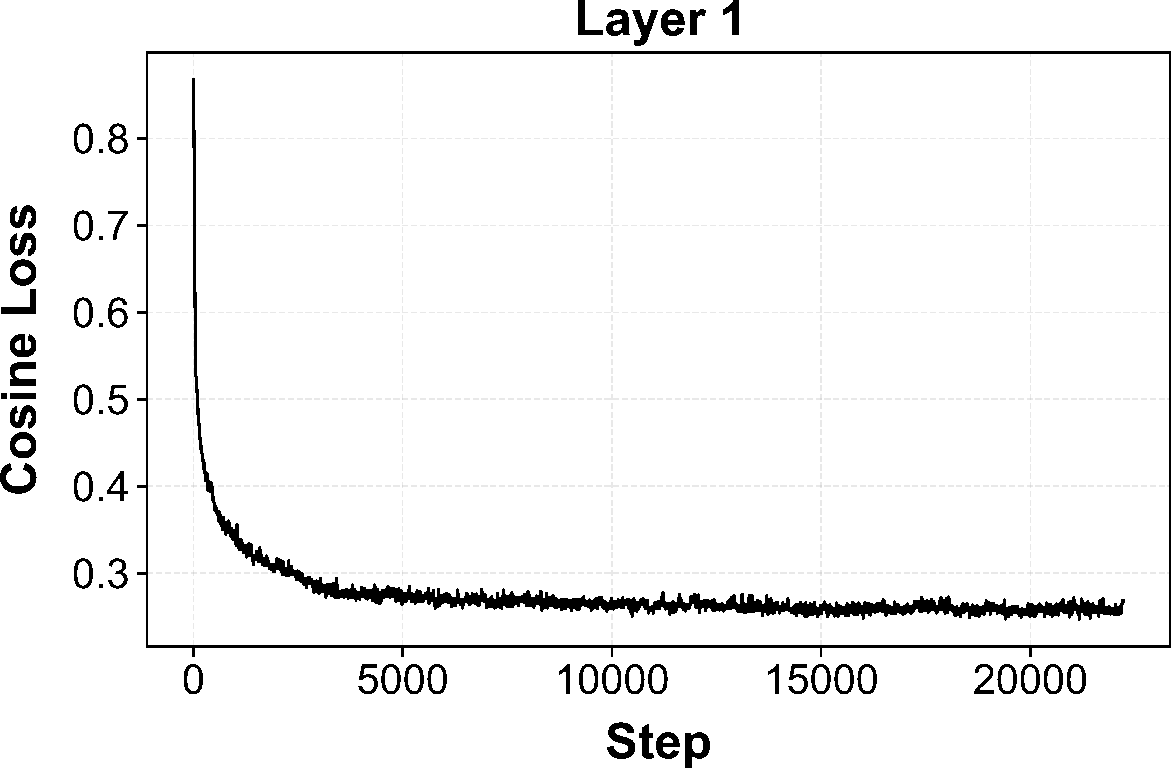}
    \end{subfigure}
    \hfill
    \begin{subfigure}[t]{0.49\linewidth}
        \centering
        \includegraphics[width=\linewidth]{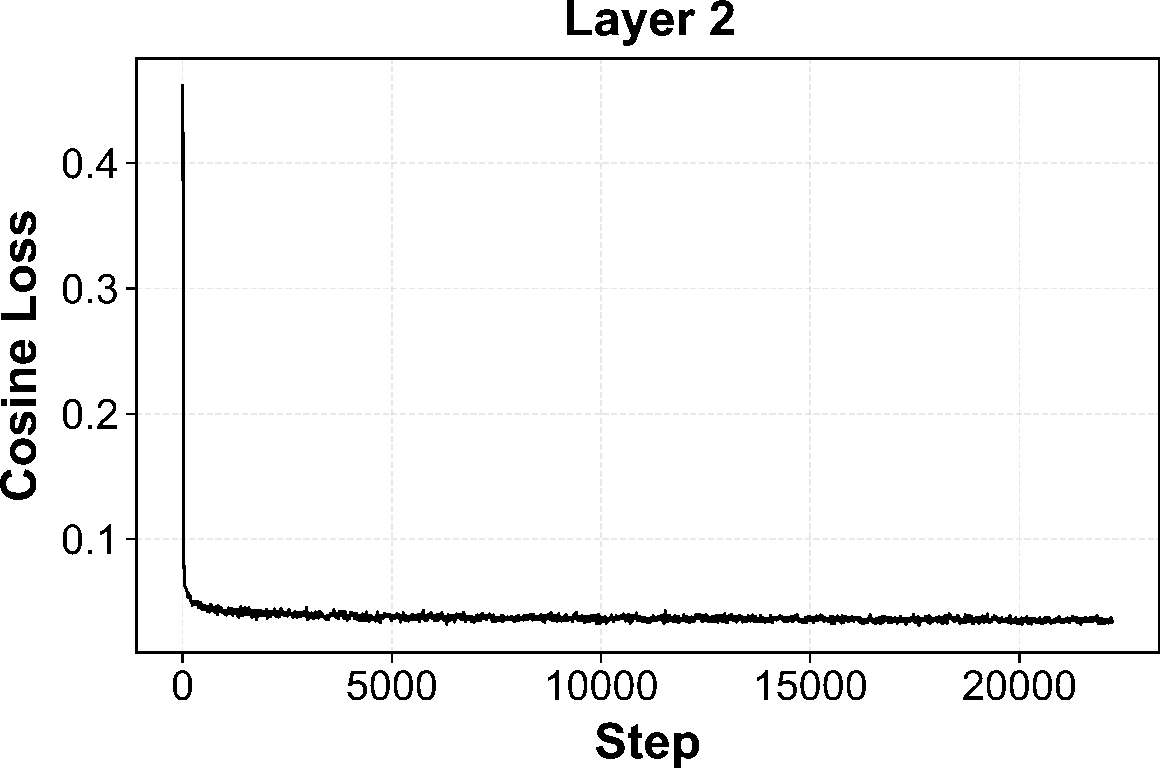}
    \end{subfigure}
    \caption{Cosine training loss for the first and second matched layers. }
    \label{fig:cosine_losses}
\end{figure}

\begin{figure}[htbp]
    \centering
    \begin{subfigure}[t]{0.49\linewidth}
        \centering
        \includegraphics[width=\linewidth]{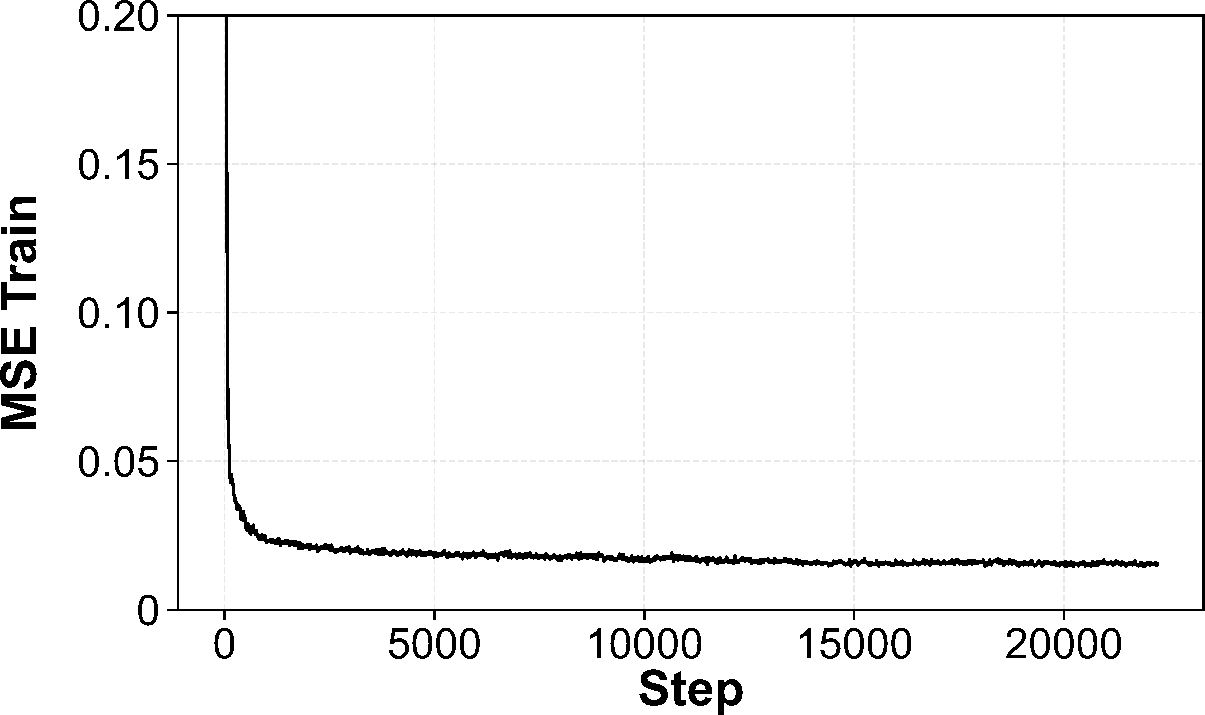}
    \end{subfigure}
    \hfill
    \begin{subfigure}[t]{0.49\linewidth}
        \centering
        \includegraphics[width=\linewidth]{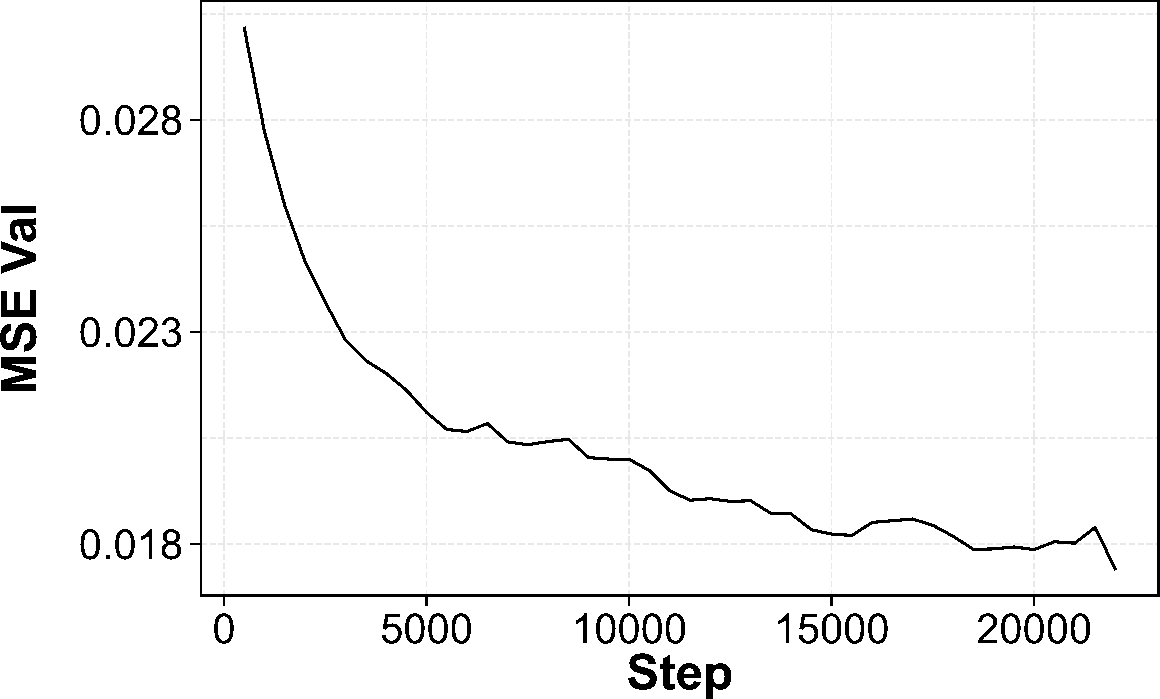}
    \end{subfigure}
    \caption{MSE Losses for training and validation data}
    \label{fig:mse_losses}
\end{figure}

\begin{figure}[htbp]
    \centering
    \includegraphics[width=0.7\linewidth]{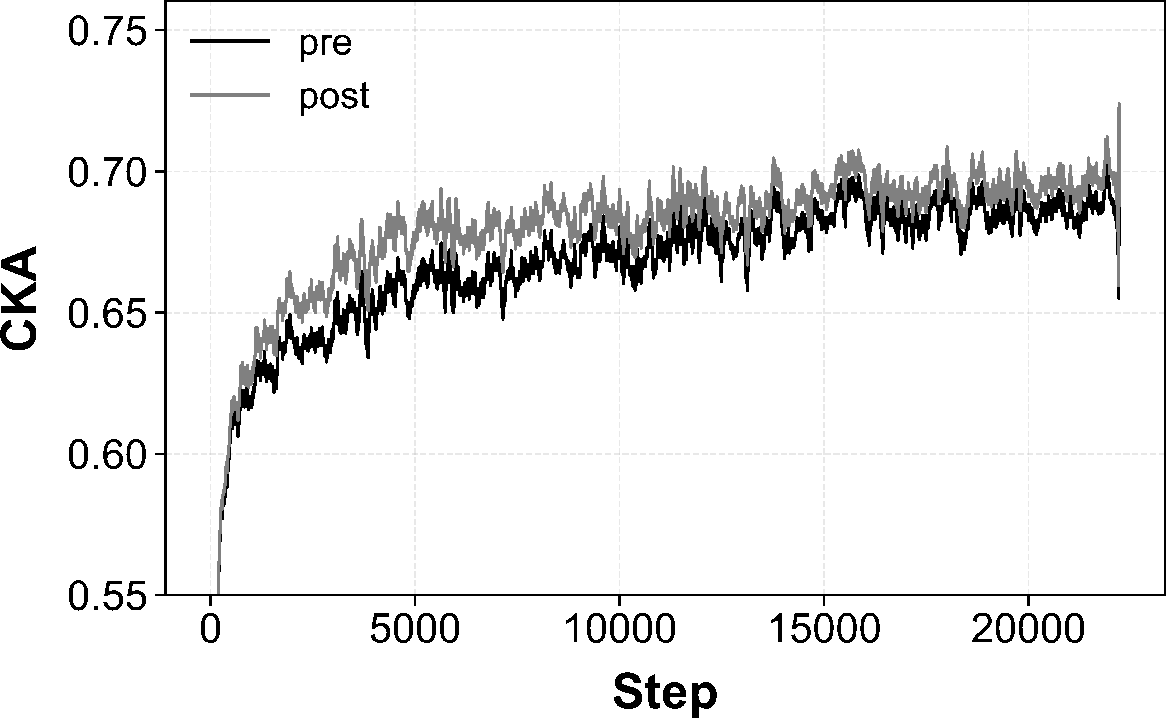}
    \caption{Central Kernel Alignment (CKA) Values Pre- and Post- Linear Projection Layer during training.}
    \label{fig:cka_plots}
\end{figure}

\begin{figure}[htbp]
    \centering
    \includegraphics[width=\linewidth]{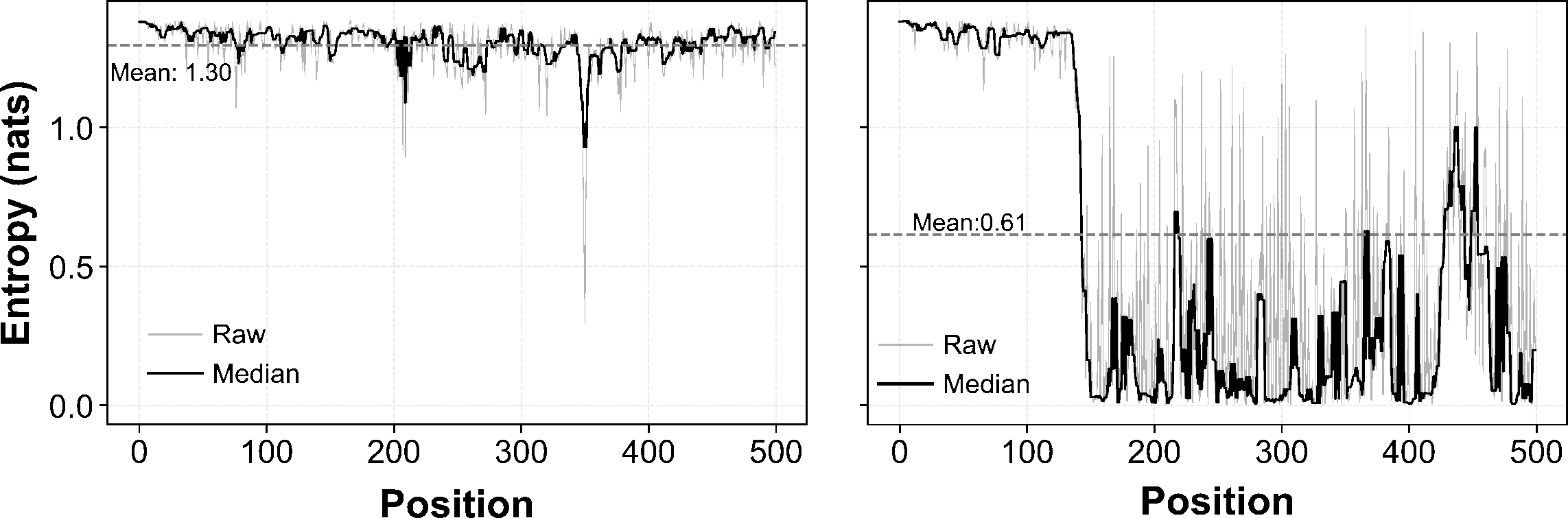}
    \caption{Entropy against position from the start of the mRNA sequence. Abrupt changes in the logit distribution and regions of high entropy make learning difficult.}
    \label{fig:entropy_noise}
\end{figure}

\begin{table}[htbp]
    \caption{Number of principal components required to reach different variance thresholds for Evo2 hidden layers.}
    \label{tab:pca}
    \begin{center}
    \begin{tabular}{ccc}
    \multicolumn{1}{c}{\bf Variance Threshold} &
    \multicolumn{1}{c}{\bf block 12} &
    \multicolumn{1}{c}{\bf norm}
    \\ \hline \\
    50\%  & 1   & 2 \\
    75\%  & 1   & 4 \\
    90\%  & 1   & 5 \\
    95\%  & 1   & 6 \\
    99\%  & 431 & 6 \\
    \end{tabular}
    \end{center}
\end{table}

\begin{figure}[htbp]
    \centering
    \includegraphics[width=\linewidth]{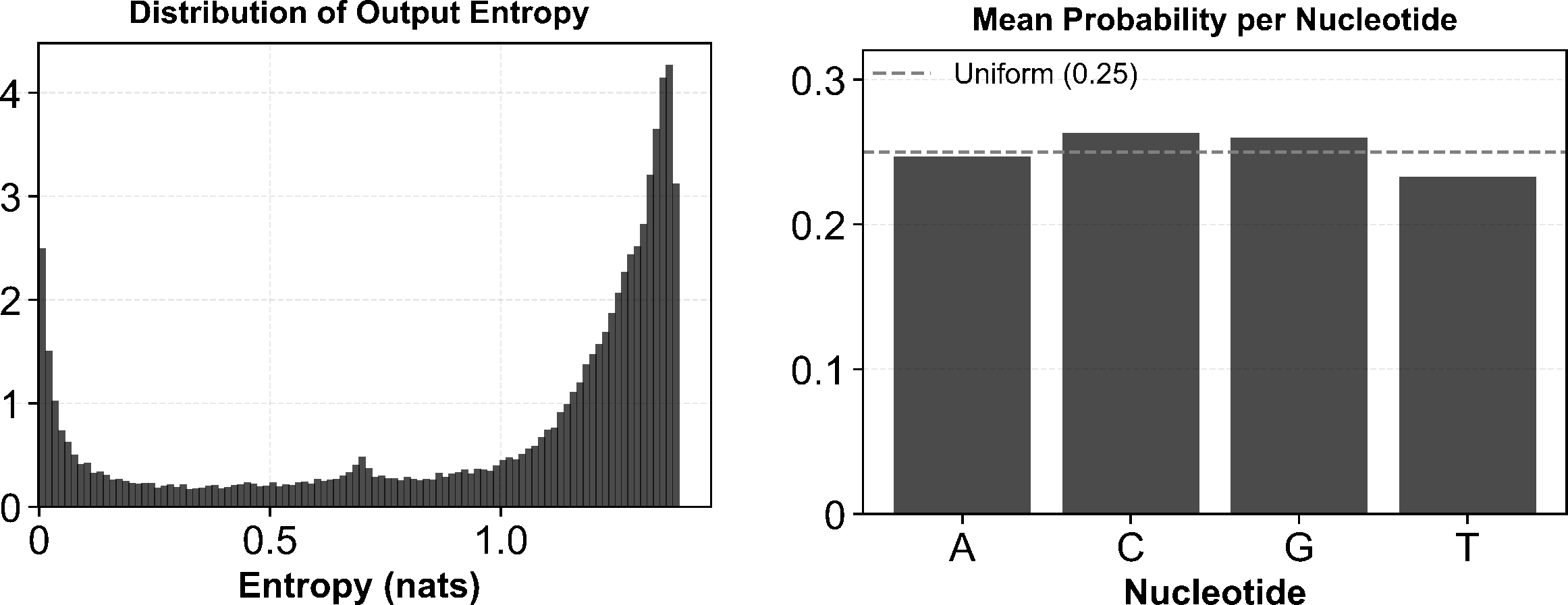}
    \caption{Summary entropy statistics across 100 mRNA sequences. The average probability per token is slightly better than uniform.}
    \label{fig:summary_ent}
\end{figure}

\begin{figure}[htbp]
    \centering
    \includegraphics[width=\linewidth]{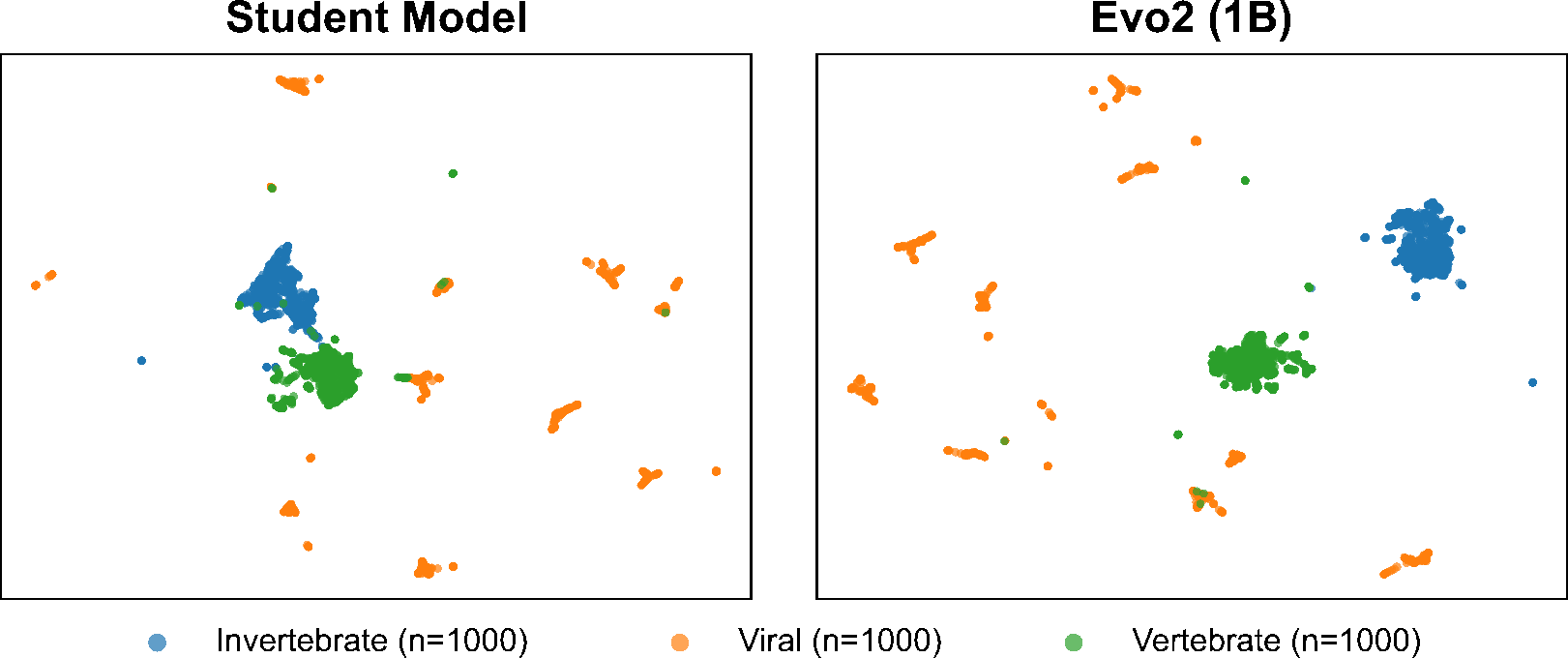}
    \caption{UMAP over Phylum for Evo2 and the student model.}
    \label{fig:umap}
\end{figure}


\end{document}